\documentclass{article}
\usepackage{amsmath}%
\usepackage{amsfonts}%
\usepackage{amssymb}%
\usepackage{graphicx}
\usepackage{listings}
\usepackage{color} 
\definecolor{mygreen}{RGB}{28,172,0} 
\definecolor{mylilas}{RGB}{170,55,241}
\begin{document}
\title{Studying Topology of Time Lines Graph leads to an alternative approach to the Newcomb's Paradox}
\author{Giuseppe Giacopelli \\ Dep. of Mathematics and Informatics, University of Palermo, Italy. \\ Institute of Biophysics, CNR, Palermo, Italy. \\ \texttt{giuseppe.giacopelli@unipa.it}, \texttt{giuseppe.giacopelli@pa.ibf.cnr.it}  }

\maketitle
\begin{abstract}
The Newcomb's paradox is one of the most known paradox in Game Theory about the Oracles. We will define the graph associated to the time lines of the Game. After this Studying its topology and using only the Expected Utility Principle we will formulate a solution of the paradox able to explain all the classical cases.
\end{abstract}

\section{Introduction}
The Newcomb's Paradox was exposed the first time in the paper written by Nozick in \cite{Nozick1969}, a paper between the Game Theory \cite{Sihlobo2013} and philosophy. The paradox is essentially a game between two players $S$ and $C$. The player $S$ plays first. he uses an oracle $\Omega$ to know what $C$ will do. $C$ plays after $S$, but he doesn't know what $S$ has chosen. $S$ has two choices $S_1$ and $S_2$. Symmetrically $C$ has two choices $C_1$ and $C_2$. Because of the fact that $S$ is the first to play, $C$'s utility function is dependent from the choice of $S$. The utilities are shown in table

\begin{center}
  \begin{tabular}{ | l | c | r | }
    \hline
     Utility & $C_1$ & $C_2$ \\ \hline
    $S_1$ & 10000 & 0 \\ \hline
    $S_2$ & 1010000 & 1000000 \\
    \hline
  \end{tabular}
\end{center}
Where the utilities are in euros. Now the problem is how can $C$ maximize his utility?

To solve the problem Nozick introduced two approaches, one based on dominant strategy and one based on the expected utility function. This approaches was used in combination to solve 6 different situations. Our aims are to create an alternative solution to the problems with the Expected Utility Function and analyze the graph of time lines to create an algorithm to test numerically the predictions. 

\section{State of the Art}
This problem was inspired (as said by the same Nozick) by Sci-Fi. The main issue of the problem is the presence of the oracle $\Omega$, which let $S$ predict the future. Then even if $S$ plays first, he knows what $C$ will do. 

The first approach is based on Expected Utility Function. It observes that $S$ forecasts future (we will see that is more complicated than this). Then if $C$ choose $C_1$ the utility must be 1000. Otherwise the utility must be 1000000. For this reasoning the best choice for $C$ is $C_2$.

The second approach is based on dominant strategy. It starts from the assumption that the oracle $\Omega$ can fail. Then the choice $C_1$ gives an utility greater of 10000 than $C_2$ in every case. In other words it is a dominant strategy.

The fact that they are logically both valid, required a mathematical investigation to classify 6 different cases. In each cases he used one of the two approaches. For the classification see \cite{Nozick1969}. We will reduce this classification to two distinct cases: The response of $S$ depends on the choice of $C$ or not. 

\section{Classic approach}
A classic approach to the problem is based on the probability. We start defining the utility table as

\begin{center}
  \begin{tabular}{ | l | c | r | }
    \hline
    Utility & $C_1$ & $C_2$ \\ \hline
    $S_1$ & $v_{11}$ & $v_{12}$ \\ \hline
    $S_2$ & $v_{21}$ & $v_{22}$ \\
    \hline
  \end{tabular}
\end{center}
where $v_{ij}$ is a positive real for every couple of indexes. We also define the probabilities $\rho_{ij}=P(S=S_i|C=C_j)$ for $i,j=1,2$. 

Now we calculate the Expected Utility Function for the choice $C_j$ as 
$$ U_j=v_{1j} \rho_{1j}+v_{2j} \rho_{2j}, \forall j=1,2. $$ 
Observing that $\rho_{1j}+\rho_{2j}=1$ for every $j$, we define $p_{1}=\rho_{11}$ and $p_{2}=\rho_{22}$. So the previous equations become
$$ U_1=v_{11} \rho_{11}+v_{21} (1-\rho_{11})=v_{21}+p_1(v_{11}-v_{21}) $$
and
$$ U_2=v_{12}(1-\rho_{22})+v_{22} \rho_{22}=v_{12}+p_2(v_{22}-v_{12}). $$
$C$ will choose choice $C_1$ if and only if
$$ v_{21}+p_1(v_{11}-v_{21})=U_1 \geq U_2=v_{12}+p_2(v_{22}-v_{12})$$
which leads to the inequality in $p_1$ and $p_2$
$$ v_{21}+p_1(v_{11}-v_{21}) \geq v_{12}+p_2(v_{22}-v_{12}).$$

\begin{figure}[htbp]
	\centering
		\includegraphics[width=0.65\textwidth]{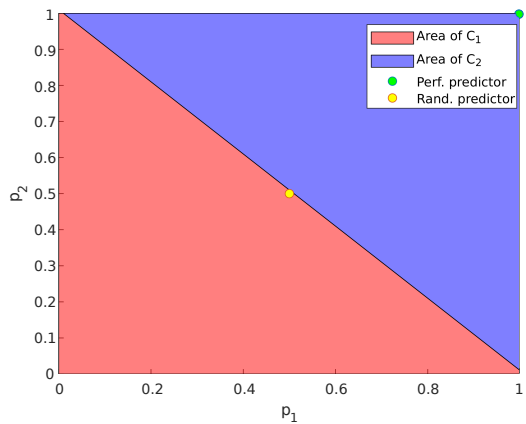}
	\caption{In figure is depicted the graphical solution of the inequality. In the Red area $C$ will choose $C_1$ and in the Blue area will choose $C_2$. Can also be seen two standard cases: The perfect predictor (green dot) and the random predictor (yellow dot).}
	\label{fig:1}
\end{figure}

With the classical values $v_{11}=10000$, $v_{12}=0$, $v_{21}=1010000$ and $v_{22}=1000000$ the inequality takes the form 
$$ p_2 \leq 1.01 - p_1 $$
which is solved graphically in Fig. \ref{fig:1}.

Three important cases are when $p_1=p_2=1$, which means that $\Omega$ never fails, when $p_1=p_2=0.5$, which means that $\Omega$ has a random behavior, and finally when $p_1+p_2=1$, when the $S$'s response is uncorrelated from the action of $C$. As can be seen the classical two approach are modelized by the Expected Utility Function, in fact the green and the yellow dots represent different strategies.

\section{Time Lines}
The classical approach does not explain how the game works, but only analyzes the results of the game. The step that we want to do is to explain the dynamics of the game with sufficient detail to simulate it. But to do it we need a new way of thinking. We firstly introduce the notion of directed graph \cite{Wilson1996}. A graph is an object made up by a set of nodes $V$ and a set of edges $E \subseteq V \times V$. Every edges has a starting node and an arriving node, so an edge can be seen as an arrow from a node to an other node. Now the time lines graph (TLG) is a graph. This graph is constructed from the principle of temporal causality. Let's make a simple example. The list 
\begin{enumerate}
	\item open the door;
	\item switch on the lights;
	\item sit down;
\end{enumerate}
is a sequence of actions (an algorithm). Now we can construct a graph with nodes the actions (so $V_{alg}=\{(1),(2),(3)\}$) and we add an edge every time there is a relation of temporal causality. For example the action $(2)$ follows the action $(1)$, then we add add an edge starting from $(1)$ to $(2)$. So the resulting graph (a time line) is
$$(1) \rightarrow (2) \rightarrow (3).$$ 
Then we define the TLG as the graph representing all the time lines of the players.

It is easy to imagine that in a good time line the resulting graph must be a line starting from the first action done and arriving to the last one. By the way this simple fact must be conjectured.

\vspace{5mm}
\textbf{Principle of time linearity:} Every time line of a player is a graph topologically equivalent to a chain (a \textit{line}) with a starting node (called \textit{starting cause}) and an ending node (called \textit{final effect}).
\vspace{5mm}

\begin{figure}[htbp]
	\centering
		\includegraphics[width=0.4\textwidth]{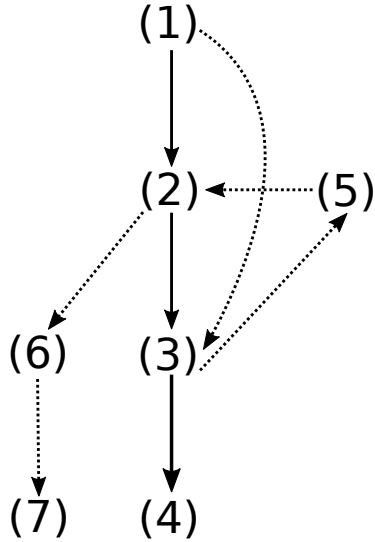}
	\caption{In figure is depicted the game time line (solid line) and the alternative time line (dashed line).}
	\label{fig:2}
\end{figure}

Now let's calculate the TLG of our game. The list of action could appear to be simple
\begin{enumerate}
	\item $S$ starts the oracle $\Omega$;
	\item $S$ gets the prevision from $\Omega$ and then it chooses;
	\item $C$ chooses;
	\item We have the result of the game.
\end{enumerate}
So the TLG should be $(1) \rightarrow (2) \rightarrow (3) \rightarrow (4)$. But it is more complex than this. When $\Omega$ finds what $C$ will do, it starts from $(1)$ arriving to $(3)$. Then it does an extra step $(5)$ where it elaborates the answer to give to $S$. Finally it comes back to $(2)$ to give the answer to $S$. The future of $(2)$ now has been perturbed by the prevision, which implies we have 2 new nodes: $(6)$ which is the new movement of $C$ (equivalent to the node $(3)$) and $(7)$ which is the new outcome of the game (equivalent to the node $(4)$). The resulting graph is depicted in Fig. \ref{fig:2}. As can be seen the TLG is not a chain, but we will see that the time line of every player it is. 

Now we are ready for the time lines of every player. The player $C$ has the simplest time line, which is 
$$(1) \rightarrow (2) \rightarrow (3) \rightarrow (4).$$
This player doesn't know what $S$ and $\Omega$ are doing. He just waits an answer from $S$. 

The time line for $S$ is a bit more complicated. $S$ knows that the oracle will perturb the future. But $S$ is not able to follow $\Omega$, then he just waits an answer from $\Omega$. So $S$'s time line is then 
$$(1) \rightarrow (2) \rightarrow (6) \rightarrow (7).$$ 

Can be observed that $(3)$ is parallel to $(6)$ and $(7)$ is parallel to $(4)$. So it seems that the game has two distinct outcomes, if the observer is $C$ the outcome is $(4)$, while if the observer is $S$ the outcome is $(7)$. To fix this problem we introduce a key concept: the entanglement between two events. We will define two events entangled if they share the same outcome. For example events $(3)$ and $(6)$ are the same event with or without the contribution of $\Omega$. But when $C$ makes his choice, he does not know what $S$ will choose, so $\Omega$ has not influences on the outcome of $(3)$ and $(6)$. Then these events must have the same outcome because they are caused by the same events. So they are entangled. We will express the entanglement with the symbol 
$$(3)\leftrightarrow (6).$$ 
Now we are ready to formulate an other principle.

\vspace{5mm}
\textbf{Principle of Retrocausation:} Every time that a time line has a singularity, the singularity split in two or more copies of the singular event entangled between them.
\vspace{5mm}

The previous principle is the one which justify the creation of the node $(6)$. But we have an other problem now. The fact that $(3)$ and $(6)$ are entangled does not guarantee us that the consequences ($(4)$ and $(7)$) are again entangled. So we make an other statement.

\vspace{5mm}
\textbf{Principle of Entanglement transmission:} The consequences of entangled events are again entangled.
\vspace{5mm}

From the previous follows that $(4) \leftrightarrow (7)$, which leads to a unique result (because of the entanglement).
Then for $\Omega$ the time line is 
$$(1) \rightarrow (3) \rightarrow (5) \rightarrow (2) \rightarrow (6) \rightarrow (7).$$ 
Let's check it:
\begin{itemize}
\item  $\Omega$ receives the input of $A$ $(1)$; 
\item Then $\Omega$ goes in future $C$'s choice to answer $S$ $(3)$; 
\item $\Omega$ elaborates the query $(5)$; 
\item $\Omega$ answers $S$ $(2)$;
\item $\Omega$ sees entangled $C$'s choice $(6)$;
\item $\Omega$ sees entangled result $(7)$.
\end{itemize}

\begin{figure}[htbp]
	\centering
		\includegraphics[width=0.4\textwidth]{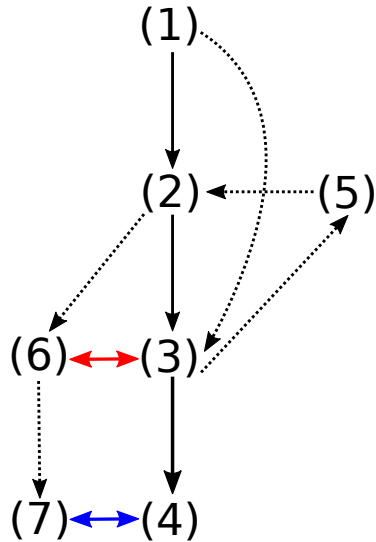}
	\caption{In figure is shown the entanglement between the nodes representing the choice of $C$ (Red) and the nodes of the result (Blue). }
	\label{fig:3}
\end{figure}

Then all the principles stated have been applied successfully (in particular the linearity principle). For a brief scheme of all the entanglements see Fig. \ref{fig:3}.

\section{Time unfolding}
We have now all the time lines of all players. Let's see them again
$$ C: (1) \rightarrow (2) \rightarrow (3) \rightarrow (4),$$
$$ S: (1) \rightarrow (2) \rightarrow (6) \rightarrow (7),$$
$$ \Omega: (1) \rightarrow (3) \rightarrow (5) \rightarrow (2) \rightarrow (6) \rightarrow (7).$$

The first idea to simulate with a computer the problem could be simulate the simplest time line, so the $C$'s one. But using this approach we encounter an insuperable obstacle. We have not an oracle. So simulating this problem in this way is impossible. The same happens for the time line of $S$. So the best way to do it is to take the weirdest way. Then we choose the time line of $\Omega$. In fact what we observe is that the event $(3)$ in his time line happens before the event $(2)$. This simple observation leads to an notable consequence. Retrocausation has caused a twist in the time line of observer $\Omega$. Then causality is not an invariant for a physical systems but can be perturbed by using an oracle. This simple observation has implications in physics but these are out of our aims. By the way appear clear that we are able to write a program which simulates the game from the frame of $\Omega$. Thanks to time unfolding caused by the twist of the time line is possible. We expose a simple MATLAB implementation.

\lstset{language=Matlab,%
    breaklines=true,%
    morekeywords={matlab2tikz},
    keywordstyle=\color{blue},%
    morekeywords=[2]{1}, keywordstyle=[2]{\color{black}},
    identifierstyle=\color{black},%
    stringstyle=\color{mylilas},
    commentstyle=\color{mygreen},%
    showstringspaces=false,
    numbers=left,%
    numberstyle={\tiny \color{black}},
    numbersep=9pt, 
    emph=[1]{for,end,break},emphstyle=[1]\color{red}, 
}

\begin{lstlisting}
% (1) S calls Omega, so the function starts
rng('shuffle')
% (3) C chooses Ci
E3=<1 or 2>;
% (5) Omega chooses the prediction
pt=p(E3);
if rand()<pt
    % keep the prevision
    E5=E3;
else
    % switch the prevision
    E5=3-E3;
end
% (2) Omega sends the prediction to S
E2=E5;
% (6) C plays again (Entangled with (3))
E6=E3;
% (7) Result
E7=v(E2,E6);
% (4) is Entangled with (7)
E4=E7;
% End
\end{lstlisting}
Where the matrix $v$ is the $2 \times 2$ utility matrix and $p$ is the vector with elements $p_1$ and $p_2$.
    
\section{Results}
Through the the MATLAB implementation has been tested the predictions of the classical theory in a numerical way. The game has been played $N=50000$ times with a parallel code. Then the obtained utilities have been averaged. The results are shown in tables. In the first test we suppose $p_1=p_2=0.5$ (Random case)

\begin{center}
  \begin{tabular}{ | l | c | r | }
    \hline
    Utility & $C_1$ & $C_2$ \\ \hline
    Theoretical & 510000 & 500000\\ \hline
    Numerical & 516480 & 496560 \\
    \hline
  \end{tabular}
\end{center}
and in the second test we suppose $p_1=p_2=1$ (Perfect predictor)

\begin{center}
  \begin{tabular}{ | l | c | r | }
    \hline
    Utility & $C_1$ & $C_2$ \\ \hline
    Theoretical & 10000 & 1000000\\ \hline
    Numerical & 10000 & 1000000\\
    \hline
  \end{tabular}
\end{center}
where in the second case we observe that the algorithm is deterministic. Then the algorithm proposed explains the two classical solutions, which are $C_1$ in the first test and $C_2$ in the second test.
\bibliographystyle{plain}
\bibliography{REF}
\end{document}